\newcommand{\minisection}[1]{\noindent {\bf #1}}

\documentclass[conference, peerreview]{IEEEtran}
\usepackage{cite}
\usepackage{amsmath,amssymb,amsfonts}
\usepackage[noend]{algorithmic}
\usepackage{algorithm}
\usepackage{algorithmic}
\usepackage{graphicx}
\usepackage{textcomp}
\usepackage{xcolor}
\usepackage{stfloats}
\usepackage{hyperref}

\def\BibTeX{{\rm B\kern-.05em{\sc i\kern-.025em b}\kern-.08em
    T\kern-.1667em\lower.7ex\hbox{E}\kern-.125emX}}
\begin{document}

\title{NTRL: Encounter Generation via Reinforcement Learning for Dynamic Difficulty Adjustment in Dungeons and Dragons
% {\footnotesize \textsuperscript{}Note: Sub-titles are not captured for https://ieeexplore.ieee.org  and
}

\author{\IEEEauthorblockN{Carlo Romeo}
\IEEEauthorblockA{University of Florence \\
carlo.romeo@unifi.it}

\and

\IEEEauthorblockN{Andrew D. Bagdanov}
\IEEEauthorblockA{University of Florence \\
andrew.bagdanov@unifi.it}
}
% \author{Anonymous}

\maketitle
\begin{abstract}
Balancing combat encounters in Dungeons \& Dragons (D\&D) is a complex task that requires Dungeon Masters (DM) to manually assess party strength, enemy composition, and dynamic player interactions while avoiding interruption of the narrative flow. In this paper we propose Encounter Generation via Reinforcement Learning (NTRL), a novel approach that automates Dynamic Difficulty Adjustment (DDA) in  D\&D via combat encounter design. By framing the problem as a contextual bandit, NTRL generates encounters based on real-time party members attributes. In comparison with classic DM heuristics, NTRL iteratively optimizes encounters to extend combat longevity (+200\%), increases damage dealt to party members, reducing post-combat hit points (-16.67\%), and raises the number of player deaths while maintaining low total party kills (TPK). The intensification of combat forces players to act wisely and engage in tactical maneuvers, even though the generated encounters guarantee high win rates (70\%).
Even in comparison with encounters designed by human Dungeon Masters, NTRL demonstrates superior performance by enhancing the strategic depth of combat while increasing difficulty in a manner that preserves overall game fairness.
Source code is available at \href{https://github.com/CarloRomeo427/NTRL.git}{github.com/CarloRomeo427/NTRL.}
\end{abstract}


\IEEEpeerreviewmaketitle

\section{Introduction}

Dungeons \& Dragons (D\&D)~\cite{dnd} is a tabletop role-playing game (TTRPG) that integrates narrative depth with strategic gameplay. Since its inception in 1974, it has evolved into a complex system in which the Dungeon Master (DM), the narrator, orchestrates the game world through their control of non-player characters (NPCs), by adjudicating rules, and by managing contingencies and events, while players assume the roles of distinct characters navigating various challenges. Gameplay is structured around turn-based encounters, where combat, diplomacy, and problem-solving are central to progression. Outcomes are determined by dice rolls, character statistics, and tactical positioning, creating a dynamic and immersive experience.

One of the most delicate responsibilities of a DM is balancing combat encounters to ensure they are neither trivial nor excessively punishing. Effective encounter design requires careful analysis of party composition, character abilities, and player strategies. Poorly tuned encounters can lead to disengagement if too easy or frustration if too difficult. The Dungeon Master’s Guide (DMG)~\cite{dmg} provides static guidelines based on challenge ratings (CRs) and experience point (XP) budgets. However, these guidelines fail to fully capture synergies and real-time player conditions.

Automatic playtesting methods have gained traction to address these limitations and relax the duties of the Dungeon Master~\cite{auto_playtest_dnd, no_player_left_behind}. Proposed methods tackle the problem from different perspectives: (i) an extensive simulation phase to estimate the optimal solution; (ii) mimicking human player behavior; or (iii) designing heuristics based on quantitative insights about difficulty scaling, player performance, and combat balance. However, the core limitation of most approaches is reliance on lengthy simulations which may disrupt the usability of these applications in real-time campaigns by breaking immersion. Dynamic Difficulty Adjustment (DDA) methods offer an alternative to adapt the game features to players ability in real-time~\cite{DDA_review}. Although the methodologies used by DDA approaches are similar to those for Automatic Playtesting, DDA moves the application time directly into the development of the game itself.
% Unlike dynamic difficulty adjustment (DDA) in digital games, D\&D encounters remain fixed once set, often necessitating mid-session DM adjustments that can disrupt immersion. Traditional playtesting in tabletop settings is impractical due to time constraints, making computational approaches crucial for refining encounter design. By leveraging AI-driven simulations, encounter balance can be dynamically optimized, ensuring engaging, strategically challenging, and narratively cohesive gameplay experiences.

In this paper we propose E\textbf{N}counter Genera\textbf{T}ion via \textbf{R}einforcement \textbf{L}earning (NTRL), a novel approach to Dynamic Difficulty Adjustment (DDA) in Dungeons \& Dragons (D\&D) through automated combat encounter design. The core innovation of NTRL lies in the application of Reinforcement Learning (RL)~\cite{sutton} to dynamically adjust game difficulty in response to real-time player conditions. Our approach departs from conventional heuristic-based methods that rely on static experience point calculations. NTRL autonomously determines optimal enemy compositions by estimating the maximum challenge a party can endure while maintaining balance and preventing overwhelming difficulty.

Unlike traditional balancing approaches, which necessitate human intervention and are both time-intensive and skill-dependent, NTRL learns to model encounter difficulty through an extensive training phase in a simulated environment. Once trained, the model generates encounters instantaneously without requiring further human input beyond specifying the current party status. To evaluate the effectiveness of our approach, we developed a web-based evaluation platform, enabling direct comparisons between NTRL-generated and human-designed encounters. Empirical results demonstrate that NTRL consistently outperforms both heuristic-based and human-created encounter designs, improving strategic depth while maintaining balance between fairness and difficulty.

The main contributions of this work are:
\begin{itemize}
    \item A practical tool designed to assist Dungeon Masters (DMs) by dynamically generating combat encounters based on real-time party attributes. This approach provides immediate and balanced encounters without requiring manual DM intervention, ensuring a seamless and engaging gameplay experience.
    \item Substantial enhancement of combat longevity (+200\%), a notable decrease in post-combat hit points (-16.67\%), while preserving high win rates (70\%).
    \item Comparative analysis against human Dungeons Masters confirms our proposed approach as an effective solution to automatically assess difficulty and promote tactical decisions by raising the level of challenge.
\end{itemize}

\section{Related Work}
In this section we review recent work from the literature most related to our contributions.

\minisection{Automatic Playtesting.} Game development is an iterative process that requires extensive testing to refine mechanics, balance difficulty, and enhance player engagement. Traditional playtesting relies on human feedback, but this approach is time-consuming and resource-intensive.
% To address these challenges, automatic playtesting techniques employ AI-driven simulations to evaluate game mechanics and procedural content generation (PCG) \cite{MISSING} dynamically produces game content, reducing manual effort. 
Automatic playtesting leverages a variety of AI-based techniques to simulate different playstyles and evaluate possible game states. Monte Carlo Tree Search (MCTS) \cite{mcts} is commonly used to model procedural personas—archetypal AI-driven player models that replicate diverse player behaviors \cite{auto_playtest_mcts}. Machine learning enables agents to learn from simulated encounters and optimize strategies \cite{auto_playtest_active_learning}. Additionally, rule-based systems provide structured heuristics to analyze game scenarios, while general AI models offer adaptability to new challenges \cite{auto_playtest_dnd}. Moreover, Genetic Algorithms \cite{ga} help adjust enemy numbers, enemy classes, and the number of encounters during the in-game day \cite{no_player_left_behind}. 

However, the proposed approaches based on automatic playtesting still require that the Dungeon Master determine the desired difficulty and also entail long simulation phases that hinder their impromptu implementation during the game session. 

\minisection{Dynamic Difficulty Adjustment (DDA).} These types of game design approaches dynamically modify game parameters in real-time by analyzing player performance or behavioral data to match player skill level and ensure a balance between engagement and challenge~\cite{DDA_review, AI_in_DDA}.  DDA has been widely used in video games across many genres, including action games, role-playing games (RPGs), and serious games for rehabilitation and education \cite{DDA_for_serious_games}.

Traditional implementations of DDA have relied on heuristic-based approaches, where predefined rules dictate how difficulty adapts based on player performance, for example adjusting enemy accuracy based on player performance \cite{DDA_review}. However, rule-based methods have limitations in their adaptability and scalability, prompting the integration of machine learning techniques into DDA. Some approaches use Bayesian models to analyze player behavior and dynamically calibrate game difficulty~\cite{DDA_using_RL}. Neural networks and Reinforcement Learning (RL) have also been explored to automate difficulty adjustments by learning from real-time player interactions \cite{Investigating_RL_for_DDA, RL_in_RPGs}. 

Reinforcement Learning \cite{sutton} has emerged as a powerful tool for DDA, allowing agents to adaptively modify game difficulty through trial-and-error learning. Unlike traditional rule-based approaches, RL models can optimize difficulty settings dynamically by maximizing a reward function that aligns with player engagement~\cite{DDA_using_RL, A_framework_for_designing_RL_in_DDA}. Several studies have implemented RL-powered DDA in action games, where RL agents adjust enemy behavior and attack patterns based on the player's reaction time and proficiency \cite{Investigating_RL_for_DDA}. Similarly, Q-learning agents have been used in RPGs to dynamically modify combat mechanics, ensuring that AI opponents provide an appropriate level of challenge without overwhelming the player \cite{RL_in_RPGs}.

Although there has been growing interest in the use of RL within Dungeons \& Dragons \cite{DnD_and_DQN}, the use of RL techniques in DDA for tabletop role-playing games is still relatively unexplored.

\begin{figure*}
\centering
\includegraphics[width=\linewidth]{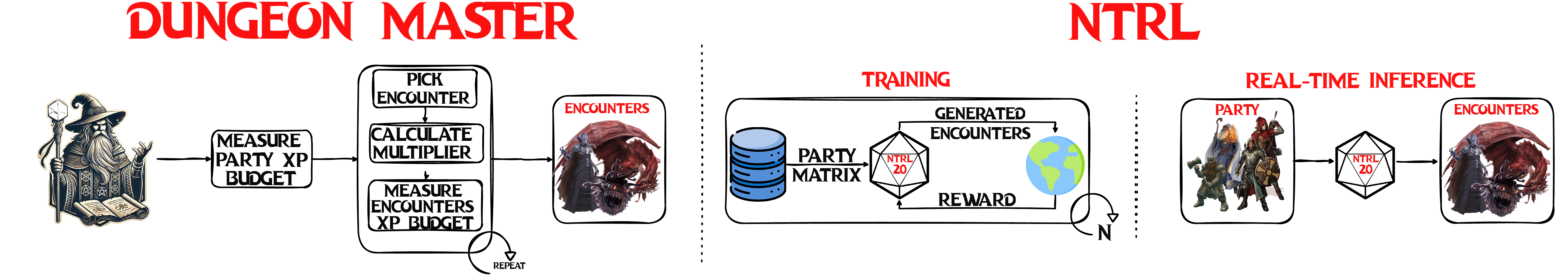}
\caption{\textbf{Human Dungeon Master versus NTRL:} A DM, based on his or her skills, must decide what the best level of difficulty is to create a stimulating event. After measuring the party XP budget, the DM must iteratively design an encounter, calculate the XP multiplier if more than one encounter has already been picked, and finally compare the encounter XP budget to that of the party to make these as close as possible. The whole process must be done (i) before the gaming session to avoid wasting time and compromising narrative; or (ii) during the game session itself, thus requiring a great effort from the DM to manage the aspects of the game altogether. On the other hand, during the training phase NTRL leverages Reinforcement Learning to iterate over a procedurally generated dataset of party compositions and interact with a simulation of the D\&D combat system, to model dynamically the most appropriate current difficulty level and optimize a reward function that promotes prolonged, tactically rich battles while minimizing the likelihood of Total Party Kills. In the inference phase, NTRL immediately generates encounters by dynamically assessing difficulty, optimizing enemy synergy, and balancing experience economy in real time. This enables the rapid creation of combat scenarios that are both engaging and strategically challenging while maintaining an appropriate level of difficulty.
}
\label{fig:teaser}
\end{figure*}

\section{NTRL: Encounter Generation via Reinforcement Learning}
In Dungeons \& Dragons (D\&D)~\cite{dnd}, the Dungeon Master (DM) is responsible for designing combat encounters that balance challenge and fairness while maintaining narrative immersion. This process is inherently complex, requiring a deep understanding of players capabilities to ensure engagement without making encounters either trivial or excessively difficult.
Despite the growing interest in Dynamic Difficulty Adjustment (DDA)~\cite{DDA_review} for tabletop role-playing games (TTRPGs), existing approaches do not specifically address adaptive difficulty in D\&D through encounter generation. Furthermore, current Automatic Playtesting~\cite{apt} solutions still necessitate DM intervention to define difficulty levels and do not provide real-time encounter generation, limiting their ability to seamlessly integrate into the ongoing narrative. This approach enables a more dynamic and personalized encounter generation process by integrating player-specific attributes into the decision-making process.

To overcome these limitations, we introduce E\textbf{N}counter Genera\textbf{T}ion via \textbf{R}einforcement \textbf{L}earning (NTRL). NTRL (Figure~\ref{fig:teaser}) is a contextual bandit~\cite{bandit} agent that is trained in a simulated environment to maximize a hand-crafted reward function that promotes long combats with high damage dealt to the party while avoiding a total annihilation of the players, causing the end of the game session. In contrast to the static guidelines provided in the Dungeon Master's Guide, which structure encounter design through a fixed resource allocation that assigns a budget for acquiring adversaries based on the desired challenge level, the number of players, and their respective levels, NTRL incorporates an embedded and comprehensive representation of the unique characteristics of each player. Moreover, to prevent overfitting on these features, we dynamically apply stochastic variations of the Hit Points for each player. 

During training, which is conducted before the application of NTRL in live game sessions, through repeated interactions with a simulation of the D\&D combat system, NTRL learns how to model difficulty on procedurally generated parties and refine its policy via the REINFORCE~\cite{sutton, reinf} algorithm. The computation and time expense are limited to the training phase, making NTRL a ready-to-go tool, as it can immediately generate of encounters at inference time. This real-time adaptation surpasses heuristic-based methods, allowing for encounters that are both well-balanced and tactically engaging without requiring extensive manual adjustments from a human DM.

% \textcolor{blue}{rephrase:
% We introduce E\textbf{N}counter Genera\textbf{T}ion via \textbf{R}einforcement \textbf{L}earning (NTRL), a novel approach to Dynamic Difficulty Adjustment (DDA) in Dungeons \& Dragons (D\&D) via combat encounters design. The core feature of NTRL is leveraging Reinforcement Learning to dynamically adapt that game difficulty on real-time player conditions, departing from official heuristic-based methods that rely on static experience point calculations. NTRL autonomously determines optimal enemy compositions by assessing the maximum challenge a party can handle while ensuring encounters remain balanced and not overwhelming. Unlike previous balancing approaches, which require human intervention, that are also time consuming and skill based, our model learns to model difficulty through an extensive training phase in a simulated environemnt, whereas after this phase the dynamic generation of encounters is immediate and does not require any human intervention expect for inputting the current party status. To validate our approach, we developed a web-based evaluation platform, enabling comparisons between human-designed and AI-generated encounters. Empirical results demonstrate that NTRL consistently outperforms heuristic and human-generated encounters, enhancing player engagement and strategic depth while maintaining fairness.}

In the following subsections we first provide a brief overview of the basic concepts of Reinforcement Learning and then introduce  the core features of the NTRL approach.

\subsection{Preliminaries}
Reinforcement Learning (RL)~\cite{sutton} is a framework for sequential decision-making in which an agent interacts with an environment to maximize cumulative rewards. It is formulated as a Markov Decision Process (MDP) $(S, A, P, R, \gamma)$, where $S$ is the state space, $A$ the action space, $P(s' | s, a)$ the transition probability, $R(s, a)$ the reward function, and $\gamma \in [0,1]$ the discount factor. The agent seeks to learn an optimal policy $\pi(a | s)$ that maximizes the expected return:
\begin{equation}
G_t = \sum_{k=0}^{\infty} \gamma^k R_{t+k}.
\end{equation}

The Contextual Bandit Problem involves decision-making in a single-step scenario in which future states are not considered~\cite{bandit}. Given a context $x \in X$, the agent selects an action $a \in A$ to maximize expected reward:
\begin{equation}
\pi(a | x) = \arg\max_{a \in A} \mathbb{E}[R | x, a].
\end{equation}
This formulation is particularly useful when decisions are independent of past interactions.

Policy Gradient~\cite{pg} methods optimize decision-making by directly updating policy parameters. The \textbf{REINFORCE}~\cite{reinf} algorithm maximizes expected rewards by adjusting policy parameters $\theta$ according to:
\begin{equation}
J(\theta) = \mathbb{E}_{\pi_{\theta}} \left[ \sum_{t=0}^{T} R_t \right].
\end{equation}
Gradient ascent is used to refine the policy:
\begin{equation}
\label{eq:grad}
\nabla_\theta J(\theta) = \mathbb{E}_{\pi_{\theta}} \left[ \sum_{t=0}^{T} \nabla_\theta \log \pi_\theta(a_t | s_t) G_t \right].
\end{equation}

\begin{algorithm}
\caption{NTRL}
\label{alg:ntrl}
\begin{algorithmic}[1]
    \STATE \textbf{Input:} Party matrix $P$, Enemy class list $E$, Policy parameters $\theta$, Max enemies $N$
    \STATE \textbf{Initialize:} Synergy vector $S$, chosen enemies $C$
    
    \FOR{$i \gets 1$ to $N$}
        \STATE Compute probabilities: $\mathbf{p} \gets \pi_\theta(P, S)$
        \STATE Sample enemy $a \sim \mathbf{p}$
        
        \IF{$a = \text{STOP}$}
            \STATE \textbf{break}
        \ENDIF
        \STATE Append $a$ to $C$ and update $S_a \gets S_a + 1$
    \ENDFOR
    
    \STATE \textbf{Simulate} encounter $(P, C)$, obtain reward $R$
    \STATE \textbf{Policy Update:} Update $\theta$ through Equation~\ref{eq:grad}
\end{algorithmic}
\end{algorithm}

\subsection{Our Approach}
In NTRL we model the combat encounter generation process as a Contextual Bandit Problem. Training is performed via simulated encounters spanning a broad range of player and NPC combinations.

\minisection{Player and NPC Representation.} Each combatant, whether a party member or an enemy, is represented by a structured JSON file containing detailed statistics. To provide a strong learning signal for NTRL, the features of each party member are encoded into a feature matrix that includes key numerical attributes such as hit points (HP), armor class (AC), core statistics, proficiency bonus, and spellcasting capabilities. Additionally, categorical variables capture saving throw proficiencies, damage resistances, spell lists, and special abilities. The party composition ranges from 3 to 8 characters, all at level 5, spanning a diverse set of predefined classes. In contrast, the enemy party consists of up to 8 opponents chosen from a pool of 26 classes, with their attributes exclusively used for simulation purposes rather than input to the NTRL model.

% REVIEWERS ASK TO DESCRIBE WHERE AND FOR THE DDA IS USED
\minisection{Dynamic HP Variations.} In real-world scenarios, the generation of encounters must also consider the current available resources of the party, such as HP, spellpoints, and potions left, because the mere class-related statistics of players cannot fully model the current capability of the party in handling a fight.
Therefore, to emulate real combat sessions in which party members may have already expended resources before combat, dynamic HP variations are introduced. Each generated party is assigned a final HP threshold from a predefined set of values: 100\%, 75\%, 50\%, 40\%, 30\%, 20\%, or 10\% of their original HP. Additionally, each character's HP is perturbed with a small noise factor between -5\% and +5\% to ensure variability in pre-battle conditions.

% To emulate real-world scenarios in which party members may have already expended resources before combat, dynamic HP variations are introduced. Each generated party is assigned a final HP threshold from a predefined set of values: 100\%, 75\%, 50\%, 40\%, 30\%, 20\%, or 10\% of their original HP. Additionally, each character's HP is perturbed with a small noise factor between -5\% and +5\% to ensure variability in pre-battle conditions.   

\minisection{REINFORCE.} Our solution (Algorithm~\ref{alg:ntrl}) employs a policy gradient method, REINFORCE, to learn optimal enemy compositions. The NTRL policy is a deep neural architecture that includes a fully connected layer with 128 neurons for each set of input features, integrating embedding for categorical features and embedding for numerical features. The embedding vectors are concatenated and sent to the final softmax layer, which produces a probability distribution on possible enemies. The action space is expanded to include a terminal action that allows the policy to generate an encounter team with a size less than the maximum. Operating within a contextual bandit framework, the agent makes a single-step decision to construct an encounter tailored to the party’s current state. 

% REVIEWERS ASK TO DESCRIBE THE SYNERGY VECTOR
The state space is augmented with a synergy vector whose length equals the number of available enemy classes. This vector is initialized to zero, and after each selection, the agent increments the entry corresponding to the chosen enemy class. Consequently, the agent’s state encodes not only the current party features but also a running count of all previously selected enemies.

% A synergy vector tracks previously selected enemies, preventing redundancy and ensuring strategic variation in encounter composition. This iterative selection process allows the policy to refine choices dynamically, accounting for the increasing difficulty imposed by multiple adversaries.

% REVIEWERS ASKED FOR MORE CLARIFICATION ABOUT THE IMPORTANCE AND THE MEANING OF EACH COMPONENT IN THE REWARD FUNCTION. THE FOLLOWING IS THE OLD SECTION THAT I REPLACE WITH A MORE EXPLICIT EXPLANATION OF WHAT WE CONSIDERED:
% \minisection{Reward Function.} The reward function evaluates encounters based on multiple combat metrics to encourage balanced and engaging battles and is the result of the combat simulation for the current party $p$ and encounter $e$ pair. Win probability ($wp$) serves as a central criterion, rewarding encounters in which the party succeeds while penalizing overwhelmingly difficult fights.
% Fight Longevity ($fl$) is incorporated to favor encounters that sustain engagement over multiple rounds. Missing HP ($mhp$) and player deaths ($dth$) are factored in to ensure that battles neither trivialize damage taken nor result in total party kills (TPK), which incur substantial penalties. Additionally, damage output ($dmg$) is considered, promoting interactions in which both teams contribute meaningfully to the combat dynamics. By integrating these components, the reward function enables NTRL to generate encounters that extend fight longevity, encourage high-damage exchanges, and maintain an engaging balance while minimizing the likelihood of a total party kill (TPK).

\minisection{Reward Function.} The reward function evaluates encounters using multiple combat metrics to ensure balanced and engaging battles, derived from simulations involving the current party $p$ and encounter $e$. Among these metrics, win probability ($wp$) serves as the foundational criterion—it is essential that encounters generated are winnable yet challenging, forming a necessary baseline before further considerations can meaningfully enhance encounter quality.

However, win probability alone cannot fully encapsulate our agent's objective. Therefore, once the baseline of a winnable encounter is established, additional metrics are integrated to refine combat dynamics. Specifically, fight longevity ($fl$) encourages encounters to last longer, maintaining player engagement throughout multiple rounds. Yet, increasing the duration alone is insufficient; encounters should not be prolonged merely by being trivially easy.
To prevent triviality, the reward function simultaneously evaluates total damage output ($dmg$) and the party’s missing HP ($mhp$). Together, these metrics ensure both sides actively participate in meaningful combat, with the party experiencing significant but manageable threats. The agent is thus incentivized to design encounters that are extended in duration while also remaining strategically engaging and demanding.
Additionally, the model allows for player deaths ($dth$) within acceptable limits, pushing the agent toward genuinely challenging encounters. However, a substantial penalty is imposed if all party members are eliminated, explicitly discouraging total party kills (TPKs):
\begin{equation}
\label{eq:rew}
% \begin{cases}
% dth = -10.000, & \text{if } dth > \text{party members} \\[6pt]
% dth = dth, & \text{otherwise}
% \end{cases}
R(p, e) = \alpha*wp + \beta*fl + \gamma*mhp + \delta*dmg + \lambda* dth
\end{equation}

% THIS IS A NEW EXPLANATION I MADE TO ANSWER THE REVIEWER ABOUT THE WEIGHT AND IMPORTANCE OF EACH METRICS
Through an extensive experimental phase, we carefully scaled each component of the reward function to reflect its intended impact and ensure proper balance during training:
\begin{itemize}
    \item \textbf{$\boldsymbol{\alpha}$:}  
    The Win Probability is scaled within the range \([0, 1000]\), ensuring that even small variations in win probability have a significant influence on the total reward. This encourages the agent to consistently prioritize generating winnable encounters.
    \item \textbf{$\boldsymbol{\beta}$ and $\boldsymbol{\gamma}$:}  
    The Fight Longevity and Missing HP metrics are scaled to operate on a comparable range. This alignment allows both to jointly influence the agent’s behavior—rewarding extended, engaging combats while penalizing encounters where the party takes little to no damage, thus avoiding trivial fights.
    \item \textbf{$\boldsymbol{\delta}$:}  
    The total damage inflicted during combat is scaled to be on the same order of magnitude as the win probability (i.e., in the thousands). This ensures that the agent also learns to balance encounters based on actual combat intensity—not only considering party size and composition, but also accounting for the pre-fight health state of the party members.
    \item \textbf{$\boldsymbol{\lambda}$:}  
    The Player's Death score is computed by summing the number of player deaths across the 100 simulations run at each training step. As a result, this metric is always a positive value and can vary significantly depending on party size and encounter difficulty. It introduces a signal of encounter lethality without completely discouraging challenging fights.
\end{itemize}

\section{Experimental Results}
\begin{figure*}[t]
\centering

% -- Legend on its own line (optional) --
\includegraphics[width=0.3\textwidth]{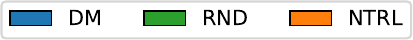}
\vspace{0.25em}

% -- First row of three plots --
\begin{minipage}[c]{0.325\textwidth}
    \centering
    \textbf{Reward}\\
    \includegraphics[width=\linewidth]{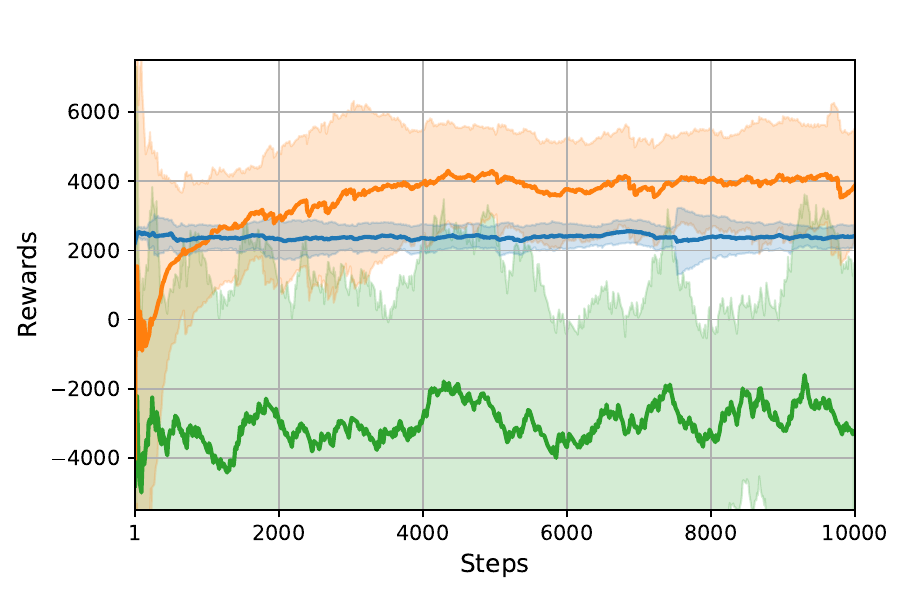}
\end{minipage}
\hfill
\begin{minipage}[c]{0.325\textwidth}
    \centering
    \textbf{Win Prob}\\
    \includegraphics[width=\linewidth]{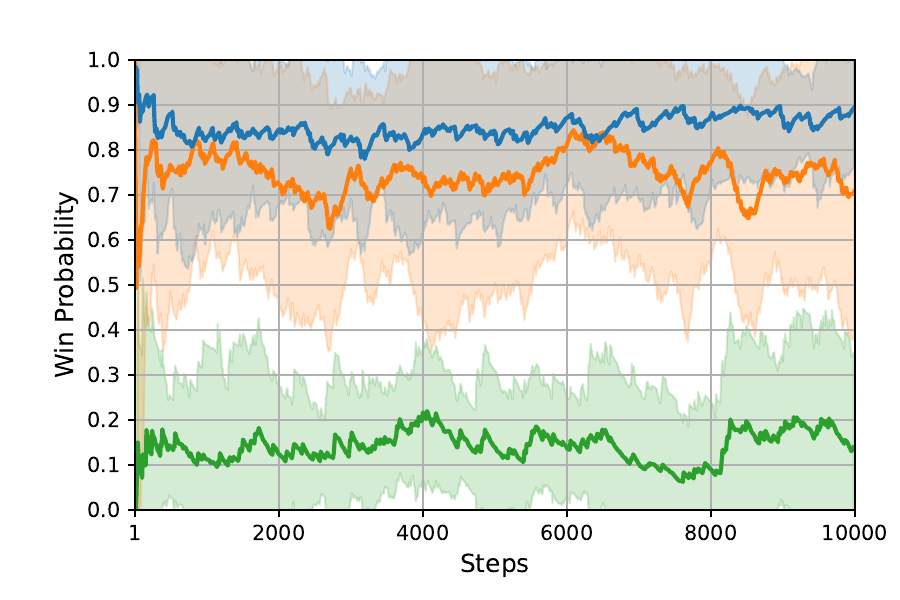}
\end{minipage}
\hfill
\begin{minipage}[c]{0.325\textwidth}
    \centering
    \textbf{Fight Longevity}\\
    \includegraphics[width=\linewidth]{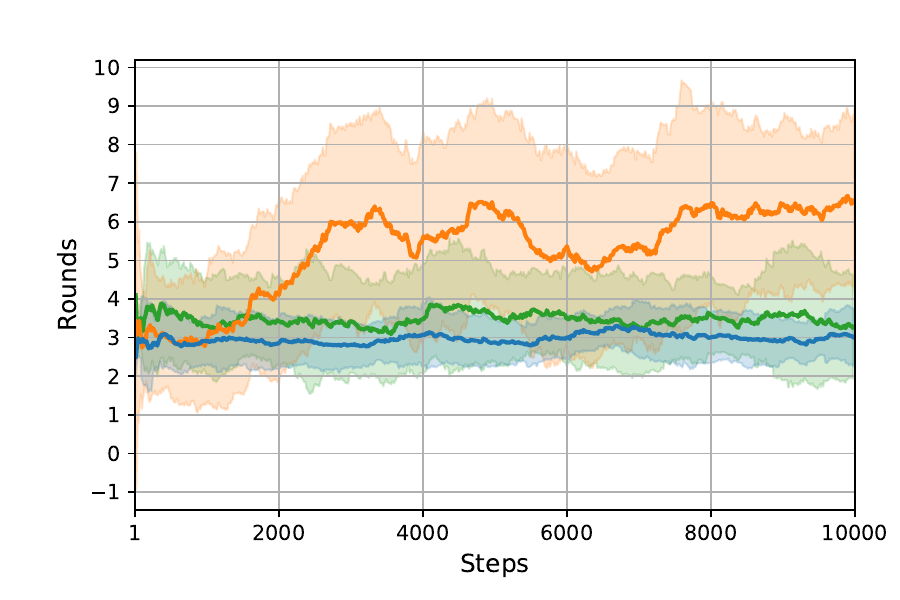}
\end{minipage}

\vspace{0.25em}

% -- Second row of three plots --
\begin{minipage}[c]{0.325\textwidth}
    \centering
    \textbf{Total Party Kills}\\
    \includegraphics[width=\linewidth]{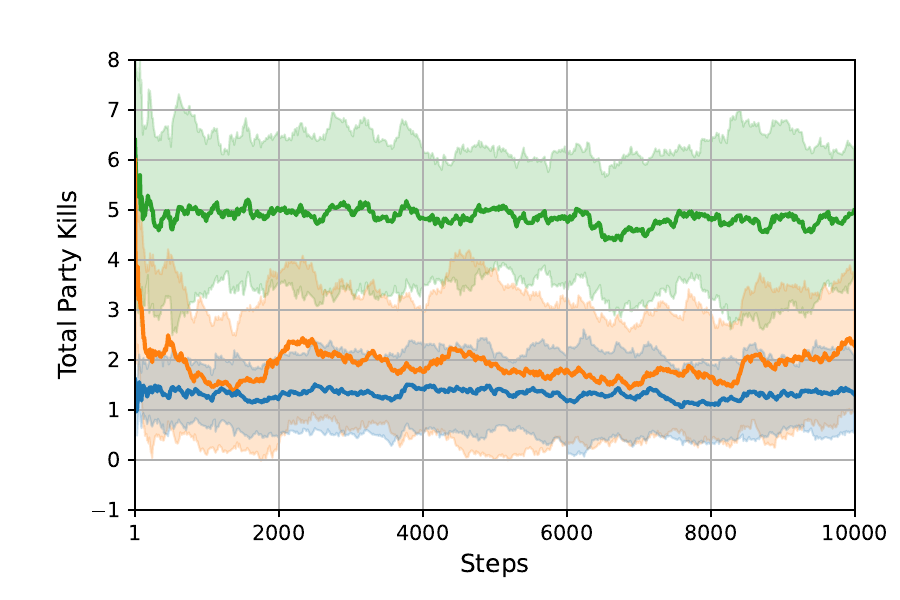}
\end{minipage}
\hfill
\begin{minipage}[c]{0.325\textwidth}
    \centering
    \textbf{Team EXP Difference}\\
    \includegraphics[width=\linewidth]{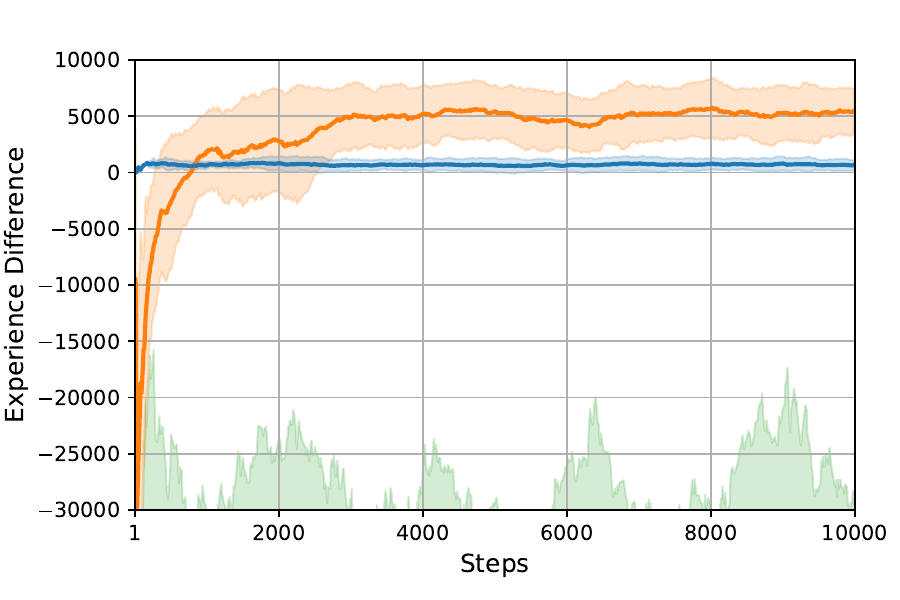}
\end{minipage}
\hfill
\begin{minipage}[c]{0.325\textwidth}
    \centering
    \textbf{Remaining Party HP}\\
    \includegraphics[width=\linewidth]{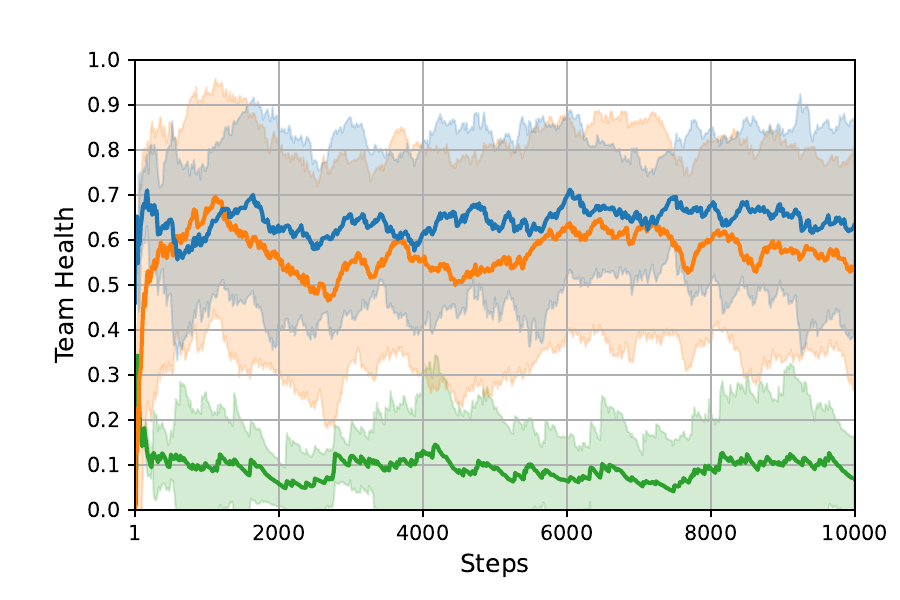}
\end{minipage}

\caption{Performance comparison of encounter generation policies. Although DM and RND policies are heuristic and do not require any training, both are plotted against steps to allow direct comparison during the evolution of the NTRL policy. The proposed NTRL policy (orange) achieves superior reward optimization and encounter balancing compared to the DM (blue) and RND (green) baselines. NTRL closely aligns with the DM policy in terms of win probability and total party kills while significantly extending fight longevity beyond both baselines, ensuring prolonged tactical engagements. It also demonstrates an emergent strategy of cost-effective XP balancing in which encounters are designed with lower XP investment yet still result in reduced remaining party hit points. This suggests that NTRL optimizes encounters to create high-damage events, forcing players into more resource-intensive combat scenarios while maintaining a balanced challenge, as evidenced by the low TPK rate, which remains similar to that of the DM policy.
}
\label{fig:wandb}
\end{figure*}
In this section, we first describe our experimental setting, including relevant metrics, the simulated environment used, and baseline policies. We then present a range of experimental results comparing the NTRL approach to baselines.
To facilitate future research, we have released the NTRL code-base here: \href{https://github.com/CarloRomeo427/NTRL.git}{github.com/CarloRomeo427/NTRL.}

\subsection{Evaluation Methodology}

To evaluate the effectiveness of our Reinforcement Learning-based encounter generation method, we compare it against two baselines: a Dungeon Master (DM) heuristic policy and a random encounter selection policy (RND). The DM policy adheres to the guidelines in the Dungeon Master's Guide (DMG), structuring encounters based on experience point (XP) budgets, while the RND policy selects enemies without any structured decision-making process. Each agent interacts with the simulated combat environment for 10,000 steps, receiving a procedurally generated party, designing a team of encounters, and finally engaging in combat simulation. To reduce variance from dice rolls, the combat simulation for each (party, encounter) pair is repeated 100 times, with results averaged. Furthermore, to address stochasticity and ensure statistically robust comparisons, the entire experiment is repeated over five random seeds.

Our evaluation is based on six key metrics:
\begin{itemize}
    \item \textbf{Reward} assesses overall agent performance using a reward function that prioritizes encounters with balanced difficulty and engagement (Equation~\ref{eq:rew}).
    \item \textbf{Win Probability} represents the proportion of simulations in which the party emerges victorious, offering insights into encounter difficulty.
    \item \textbf{Fight Longevity} measures the average number of rounds per combat, reflecting encounter depth and pacing.
    \item \textbf{Total Party Kills (TPK)} tracks instances where all party members are eliminated, indicating difficulty fairness.
    \item \textbf{Team XP Difference} compares the total experience points of the enemy team against the party’s XP budget, revealing each policy’s approach to encounter design.
    \item \textbf{Remaining Party HP} records the party’s post-combat hit points percentage, distinguishing between trivial and overly punishing encounters.
\end{itemize}
In Figure~\ref{fig:wandb}, we provide a comparative analysis of our approach against baseline policies to evaluate the efficacy of each approach in designing encounters.

\subsection{Simulated Environment}
% REVIEWERS ASK TO MAKE THE DESCRIPTION OF THE SIMULATOR MORE DETAILED, CLARIFYING THAT IT IS AN OPEN-SOURCE SIMULATOR AND POINTING OUT ITS FUNCTIONING ALSO COMPARED TO REAL-WORLD HUMAN FIGHTS
% The simulated combat environment is an open-source structured turn-based system capturing key strategic elements of live Dungeons \& Dragons (D\&D) battles~\cite{dndsim}. It fully emulates combat for both the player party and enemy combatants, with all actions autonomously controlled. Initiative-based sequencing determines action order, ensuring dynamic and responsive encounters. Each combatant is defined by attributes such as hit points (HP), initiative, status effects, class-specific abilities, and core statistics. The system tracks combat events in real-time, facilitating accurate combat reproduction and detailed post-battle analysis. 

% Combat proceeds in turns, with characters taking actions in descending initiative order until one team is entirely defeated.  Actions include melee and ranged attacks, spellcasting, defensive maneuvers, and status effect applications, with an end-of-turn phase resolving temporary conditions and buffs. The environment implements a dynamic hit points-state model, distinguishing between active, unconscious, and deceased combatants. When HP reaches zero, player characters enter a probabilistic death-saving phase, while AI-controlled enemies are removed immediately. Summoned creatures disappear upon depletion, and ongoing effects such as exhaustion, haste, and buffs persist throughout combat.

The simulated combat environment is an open-source, structured turn-based system capturing key strategic elements of live Dungeons \& Dragons (D\&D) battles~\cite{dndsim}. It fully emulates combat for both the player party and enemy combatants, with all actions autonomously controlled. Initiative-based sequencing determines action order, ensuring dynamic and responsive encounters. Each combatant, both from party members and enemies, is defined by attributes such as hit points (HP), initiative, status effects, class-specific abilities, and core statistics. The system tracks combat events in real-time, facilitating accurate combat reproduction and detailed post-battle analysis.

Combat proceeds in turns, with characters taking actions in descending initiative order until one team is entirely defeated. Each combatant is controlled by an AI that briefly assesses the current combat state, assigns a simple utility score to every legal action, and performs the option with the highest positive score, repeating until no worthwhile actions remain. Targets are chosen with the same scoring logic, and a touch of randomness keeps behaviour varied. All decisions rely on fixed heuristics, so the agents do not learn or adapt between battles. Actions include melee and ranged attacks, spellcasting, defensive maneuvers, and status effect applications. The environment implements a dynamic hit points-state model, distinguishing between active, unconscious, and deceased combatants. When HP reaches zero, player characters enter a probabilistic death-saving phase, while enemies are removed immediately. Summoned creatures disappear upon depletion, and ongoing effects such as exhaustion, haste, and temporary enhancements persist throughout combat.

Despite the effectiveness of the simulator in imitating real combat sessions, we must always consider that there will be a gap between the simulated environment and reality due to the decisions made by human players that will not always be optimal and victory-oriented.

\subsection{Dungeon Master Strategy}
The guidelines proposed in the Dungeon Master's Guide (DMG) constitute a static approach to design encounters, which is also the core caveat of this approach. Given the current party of player characters, each one identified with a level of experience, and given the desired level of challenge, which has to be properly selected by the DM, the DMG offers a table to calculate the total experience budget of the party. Once this budget is known, the DM must match this value by carefully designing a new encounter step by step, typically from the Monster Manual~\cite{monsters}, where each monster is identified with an experience cost.
Moreover, for each new encounter chosen, the DM must also consider the XP multiplier, which serves to scale difficulty with enemy count. While a single enemy retains its base XP, additional foes receive multipliers to reflect their cumulative impact, ensuring larger groups remain proportionally challenging. The DM policy we designed constructs the opponent team by evaluating all possible enemy combinations and their adjusted XP totals to match the party XP budget. 

% The result is encounters that maintain balance, avoiding trivial engagements or excessively lethal battles.
% Despite its structured framework, the  DMG’s approach based on balancing XP remains static, lacking the capacity to dynamically adjust difficulty based on real-time combat performance, a key limitation compared to adaptive AI-driven methods.

\subsection{Results}
In Figure~\ref{fig:wandb} we compare the results obtained by NTRL with the baseline agents (Heuristic DM and RND) defined above. Note that the DM and RND policies are static and do not change during the training phases. However, both are plotted against steps to serve as a direct comparison during the evolution of the NTRL policy. The plots in Figure~\ref{fig:wandb} allow comparison of NTRL with these static baselines as it converges to its final policy.

\minisection{Reward.} The NTRL policy surpasses the DM approach by optimizing encounters for engagement and challenge. Unlike the RND policy, which consistently fails to structure balanced fights, NTRL exhibits a learning trend that outperforms heuristic-based designs and achieves significantly higher reward.
%REVIEWERS THOUGHT THAT NTRL GETS MORE REWARDS JUST BY EXTENDING THE LONGEVITY OF THE FIGHTS
Note that the performance of NTRL comes not from a trivial extension of the Fight Longevity metric. Rather, the increase in reward is strongly correlated with the total damage inflicted during combat, which in turn has a direct impact on both the Win Probability and the Missing HP of the party.

\minisection{Win Probability.} While the RND policy generates excessively difficult encounters, reducing party survivability, the DM and NTRL policies maintain a stable win probability close to 0.9. NTRL initially fluctuates but ultimately converges to a balanced state, producing winnable yet challenging encounters. 

\minisection{Fight Longevity.} The duration of combats highlights structural differences in encounter balance. The DM and RND policies yield short engagements of about 3 rounds, with the former ensuring balance and the latter resulting in abrupt battles. NTRL, however, extends fights to an average of 7 rounds, fostering sustained tactical engagement.

\minisection{Total Party Kills (TPK).} The RND policy results in frequent party wipes, committing on average 5 TPKs per training step, emphasizing its poor balancing. The DM policy maintains a low TPK rate of approximately 1 per training step, ensuring fairness. NTRL closely follows this trend, achieving effective difficulty balancing without excessively punishing the party.

\minisection{Team XP Difference.} While the DM policy strictly follows XP balancing rules, the RND policy excessively inflates XP allocations. NTRL, however, consistently under-spends XP, designing encounters with 5000 fewer XP points than the DM approach. This emergent behavior suggests that the RL model autonomously learned an alternative cost-effective balancing strategy.

\minisection{Remaining Party HP.} Post-combat hit point retention reflects encounter difficulty. The DM policy ensures an average party health of around 70\%, whereas NTRL reduces it to approximately 60\%, prompting higher resource expenditure while maintaining fairness. This indicates that NTRL fosters strategic decision-making and resource management more effectively than conventional DM balancing.
\begin{figure*}[!t]
    \centering

    %---------------- LEGEND ----------------%
    % Wrap the legend in a box matching the figure width, then center the PDF
    \makebox[\linewidth]{%
        \includegraphics[width=0.5\textwidth]{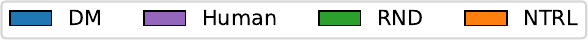}
    }%
    
    \vspace{0.5em}  % Adjust vertical space between legend and the first row of plots

    %-------------- ROW 1 --------------%
    \begin{minipage}[t]{0.25\textwidth}
        \centering
        \textbf{Reward}\\
        \includegraphics[width=\linewidth, height=4cm, keepaspectratio]{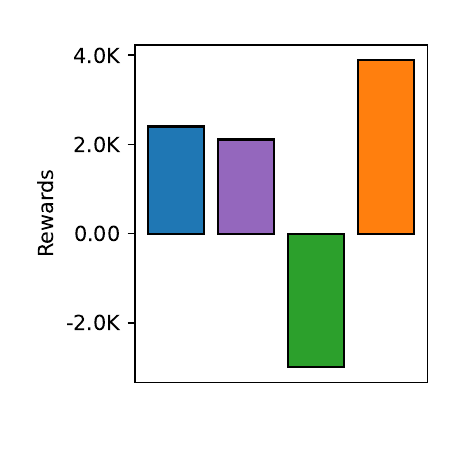}
    \end{minipage}
    \hspace{0.5em}
    \begin{minipage}[t]{0.25\textwidth}
        \centering
        \textbf{Win Prob}\\
        \includegraphics[width=\linewidth, height=4cm, keepaspectratio]{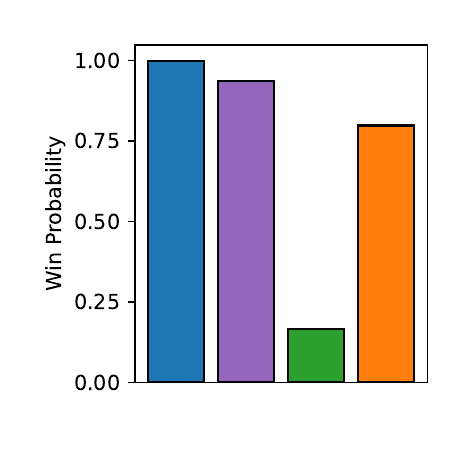}
    \end{minipage}
    \hspace{0.5em}
    \begin{minipage}[t]{0.25\textwidth}
        \centering
        \textbf{Fight Longevity}\\
        \includegraphics[width=\linewidth, height=4cm, keepaspectratio]{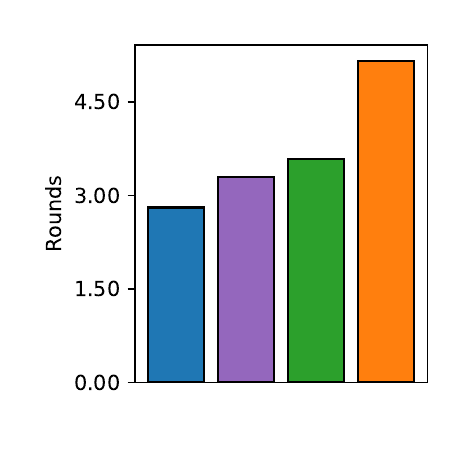}
    \end{minipage}

    \vspace{0.5em}  % Adjust vertical space between row 1 and row 2

    %-------------- ROW 2 --------------%
    \begin{minipage}[t]{0.25\textwidth}
        \centering
        \textbf{Total Party Kills}\\
        \includegraphics[width=\linewidth, height=4cm, keepaspectratio]{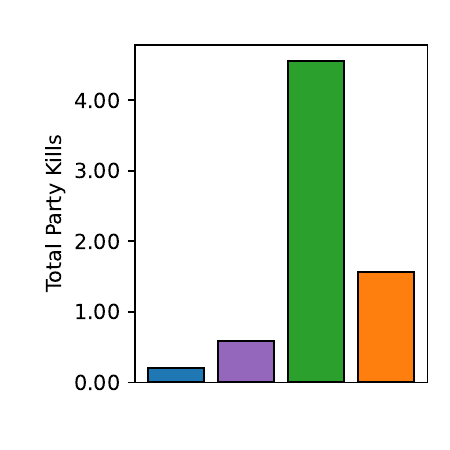}
    \end{minipage}
    \hspace{0.5em}
    \begin{minipage}[t]{0.25\textwidth}
        \centering
        \textbf{Team XP Difference}\\
        \includegraphics[width=\linewidth, height=4cm, keepaspectratio]{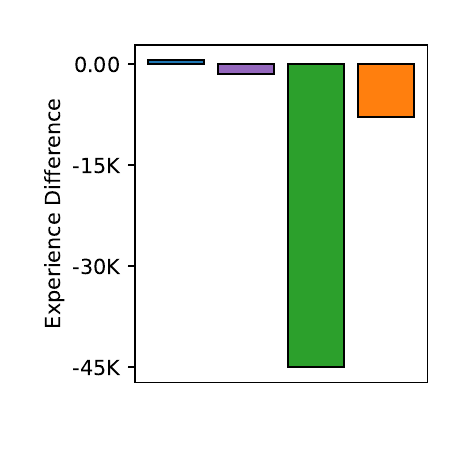}
    \end{minipage}
    \hspace{0.5em}
    \begin{minipage}[t]{0.25\textwidth}
        \centering
        \textbf{Remaining Party HP}\\
        \includegraphics[width=\linewidth, height=4cm, keepaspectratio]{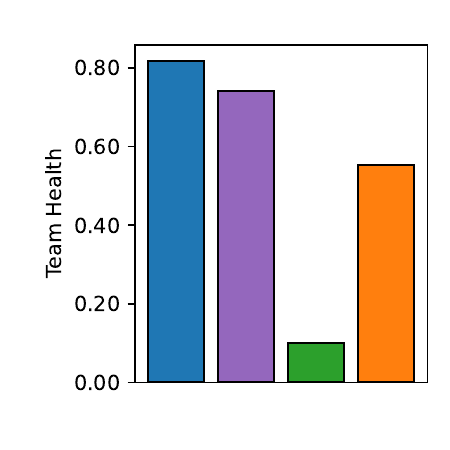}
    \end{minipage}

    \caption{\textbf{Comparison of NTRL against Human DM and heuristic DM:} The Reinforcement Learning-based NTRL policy demonstrates superior optimization of engagement and challenge compared to human Dungeon Masters (Human), heuristic Dungeon Master (DM), and random baseline (RND) methods. NTRL achieves a high win probability of 80\% while extending fight longevity beyond all baselines and maximizing rewards collection, ensuring prolonged tactical engagements. Unlike previous experiments, NTRL allocates more XP in encounter design but maintains a low total party kill rate, emphasizing strategic difficulty balancing. However, this increased XP expenditure reduces post-battle party hit points to an average of 60\%, highlighting a tradeoff between encounter challenge and survivability. The observed inversion in XP allocation trends suggests that NTRL adapts its balancing approach dynamically, but it tends to overestimate the fighting capacity of the party when no HP reduction is applied to the original HP values of each party member.
}
    \label{fig:human}
\end{figure*}

\section{Comparing NTRL with Human DMs}

To compare how NTRL and Dungeon Masters (DMs) balance encounters, we developed a web-based platform where participants design encounters for predefined, level 5 adventuring parties. This real-world data serves as a baseline for assessing Reinforcement Learning-based encounter generation under structured constraints aligned with the Dungeon Master's Guide (DMG). Each session presents a randomly generated party, and users select up to eight enemies that match the party XP budget. 
% REVIEWERS ASKED TO MAKE IT CLEAR IF THE STUDY ON HUMAN WAS SUPERVISED BY AN IRB, HOW WE INVITED THE PARTICIPANTS AND WHETHER WE SAVED EXTRA DATA FROM THE USER.
This study was not subject to oversight by an Institutional Review Board (IRB), as it did not involve the collection of any personally identifiable information beyond what was strictly necessary for the research objectives. All human submissions were fully anonymized, and no personal data were solicited from participants. The only data collected pertained to performance metrics derived from combat simulations between predefined player parties and encounters selected by participating human Dungeon Masters. Recruitment of participants was conducted via prominent online communities dedicated to Dungeons \& Dragons, including specialized Discord\cite{discord} servers and relevant Reddit\cite{reddit} forums.

Given the complexity of the task for non-expert DMs, and most of all to promote user participation, no dynamic variation of the HP is applied to the members of the party. 
Users must submit three unique encounters before evaluation through 100 combat simulations, recording metrics such as win probability, fight longevity, total party kills (TPK), and remaining party hit points. NTRL is evaluated against human DM-designed encounters (Human) and baseline methods (RND and DM). To provide an unbiased assessment of training performance, we aggregate the average performance of NTRL by using the final checkpoint from each seed. In Figure~\ref{fig:human}, we report for each evaluation metric the average results over the 86 user submissions collected.

NTRL outperforms the baselines across most of the key metrics, while maintaining a Win Probability of 75\%. Our approach again is best at promoting fight longevity with almost 5 rounds on average, whereas the DM heuristic scores even less than Human DM suggestions. The encounters proposed by NTRL depleted the remaining HP of the party even more, with a final score of around 60\% while always keeping the TPK at low values. Surprisingly, in this scenario where no dynamic reduction of HP is performed, the NTRL agent shows an inversion of the trend identified in the previous experiments: the NTRL agent prefers spending more XP in balancing the party XP budget. Therefore, the resulting Team Experience Difference average is negative. This behavior likely due to the choice of thresholds for HP reduction, where the distribution of values is skewed towards low values rather than the full HP solution as in this case.

\section{Conclusions} 

In this work we proposed Encounter Generation via Reinforcement Learning (NTRL), a novel approach for Dynamic Difficulty Adjustment in Dungeons \& Dragons (D\&D) combat encounter generation. By dynamically adjusting combat difficulty based on real-time player conditions via Reinforcement Learning, NTRL removes the reliance on static heuristics and manual adjustments, allowing for an efficient and adaptive experience. Empirical results from training simulations show that NTRL, compared to the DM's heuristic policy, prolongs combat longevity by 200\%, reduces post-fight hit points by 16.67\%, and increases tactical engagement, all while maintaining a 70\% Win Probability. These results indicate that the system effectively balances encounters so that they remain challenging without overwhelming players. Furthermore, an additional comparative evaluation conducted against human Dungeon Master designed encounters confirms that NTRL not only meets, but also exceeds traditional Dungeon Master heuristic approaches in terms of strategic depth and fairness.
% REVIEWER ASK FOR LIMITATIONS
% Beyond its technical contributions, NTRL serves as a practical tool for Dungeon Masters, streamlining encounter preparation while preserving the fluidity of in-game narratives. Future work will focus on extending character aspects, such as ammunition and class-specific characteristics, in input to the NTRL agent in order to design more complex strategies to properly challenge players, and on introducing filtering mechanisms to select which monsters are included in the encounter generation process.
Beyond its technical contributions, NTRL establishes itself as a practical tool for Dungeon Masters by streamlining encounter preparation while maintaining the fluidity of in-game narratives. However, its deployment in real-world live gaming sessions remains limited. This limitation stems primarily from constraints in the training simulator, which offers restricted customization of party members—such as fixed character levels and a limited selection of classes. Additionally, the pool of available enemies represents only a small subset of the full range of monsters found in the game.
Future work will focus on enhancing the representation of player characters by incorporating additional features, such as ammunition and class-specific attributes, into the NTRL agent's input. This will enable the design of more sophisticated and strategically diverse encounters. Furthermore, we aim to expand the encounter generation process to include filters for campaign-relevant monsters and, more importantly, to support full-day in-game encounter planning. These extensions are intended to provide comprehensive assistance to Dungeon Masters in shaping narrative progression throughout an entire session.

\bibliographystyle{IEEEtran}
\bibliography{main}

\end{document}